\documentclass[11pt]{article}
\usepackage[margin=1in]{geometry}
\usepackage{booktabs}
\usepackage{amsmath,amssymb}
\usepackage{graphicx}
\usepackage{caption}
\usepackage{xcolor}
\usepackage{hyperref}
\usepackage{enumitem}
\usepackage{tikz}
\usetikzlibrary{positioning,arrows.meta,fit,backgrounds}
\usepackage[round]{natbib}

\title{Model of Models: When Does Emitting a Specialist\\
Beat Attending, Adapting, or Tuning?\\[0.3em]
\large\normalfont A Controlled Cross-Domain Study of Amortized Weight Generation}
\author{John C. Howell\thanks{Nineonefour is the operating name of
Project[LNCLN]. Correspondence: \texttt{jc@nineonefour.org}.}\\
\small Nineonefour}
\date{\today}

\begin{document}
\maketitle

\begin{abstract}
Given a task described by a few examples, how should a model be specialized to it?
Four mechanisms are available---zero-shot, in-context attention, test-time gradient
adaptation, and emitting specialist weights from a hypernetwork---yet the operating
regime of the last is rarely mapped. We run the identical four-way comparison across
six tasks spanning regression, generation, language modeling, reinforcement learning,
and clinical and genomic classification, holding the specialist, the context, and
(where we can) the training budget fixed. The clearest wins for emission are about
\emph{cost at matched quality}: it ties the state-of-the-art amortized tabular model
(TabPFN) on clinical few-shot classification while emitting a \emph{reusable}
specialist instead of re-attending the support set per query, and reaches
noise-floor shape generation with a $132$-float per-instance program. On few-shot
sinusoid regression it is $2$--$3$ orders of magnitude below MAML at \emph{zero}
test-time gradient steps---a margin that narrows to $\sim$$30\times$ but persists once
training budgets are equalized. Emission cannot match in-context attention on
high-dimensional sequence modeling: under matched-budget pre-training a one-pass
adapter recovers only a minority of the in-context gain ($14.0\pm0.9\%$ at $5$M,
$11.2\pm0.5\%$ at $15$M), and a LoRA-rank sweep shows this shortfall is a partial
capacity limit---capture climbs from $5\%$ to $21\%$ as rank grows but plateaus far
below full recovery. Mechanism
ablations confirm the emitted specialist is genuinely task-conditioned, not a
memorized prior; and, more speculatively, emitted specialists compose in weight
space---interpolating two of them tracks the corresponding blend of their functions.
We close with a falsifiable thesis, operationalized through a per-task resolution
measure, bounding when each conditioning mechanism should be preferred.
\end{abstract}

\section{Introduction}

A recurring problem in machine learning is to specialize a general model to a
specific task that is described only by a handful of examples: a few labeled
patients, a few input--output grids, a few transitions from a new environment. Four
mechanisms are in common use. \emph{Zero-shot} ignores the examples. \emph{In-context}
learning prepends them and lets attention condition the forward pass. \emph{Gradient
adaptation}---MAML and ordinary fine-tuning---takes a few optimization steps on the
examples. The fourth, which we study here, \emph{emits} the parameters of a small
task-specific specialist network in a single forward pass of a hypernetwork. The
mechanism is not new: hypernetworks, conditional neural processes, and
context-to-LoRA generators all instantiate it. What is missing is a map of
\emph{when} it should be preferred over the other three.

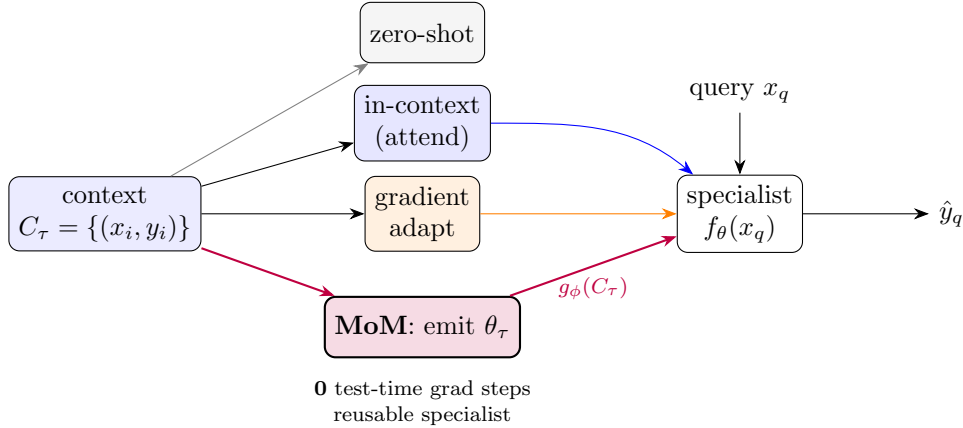
\begin{figure}[t]\centering
\begin{tikzpicture}[
  font=\small,
  box/.style={draw,rounded corners,minimum height=8mm,minimum width=15mm,align=center},
  ctx/.style={draw,fill=blue!8,rounded corners,minimum height=7mm,align=center},
  >={Stealth[length=2mm]}]
  \node[ctx] (c) at (0,0) {context\\$C_\tau=\{(x_i,y_i)\}$};
  \node[box,fill=gray!8] (zs) at (4.2,2.4) {zero-shot};
  \node[box,fill=blue!10] (ic) at (4.2,1.2) {in-context\\(attend)};
  \node[box,fill=orange!12] (ga) at (4.2,0.0) {gradient\\adapt};
  \node[box,fill=purple!14,thick] (mom) at (4.2,-1.5) {\textbf{MoM}: emit $\theta_\tau$};
  \node[box] (sp) at (8.4,-0.0) {specialist\\$f_{\theta}(x_q)$};
  \node (q) at (8.4,1.6) {query $x_q$};
  \node (yhat) at (11.2,-0.0) {$\hat y_q$};
  \draw[->,gray] (c) -- (zs);
  \draw[->] (c) -- (ic);
  \draw[->] (c) -- (ga);
  \draw[->,thick,purple] (c) -- (mom);
  \draw[->,thick,purple] (mom) -- node[below,font=\scriptsize]{$g_\phi(C_\tau)$} (sp);
  \draw[->,orange] (ga) -- (sp);
  \draw[->,blue] (ic) to[out=0,in=140] (sp);
  \draw[->] (q) -- (sp);
  \draw[->] (sp) -- (yhat);
  \node[font=\scriptsize,align=center] at (4.2,-2.5)
    {\textbf{0} test-time grad steps\\reusable specialist};
\end{tikzpicture}
\caption{The four conditioning mechanisms over a shared specialist $f_\theta$ and
query $x_q$. Zero-shot ignores the context; in-context attends to it every query;
gradient adaptation takes SGD steps on it; \textbf{MoM} emits the specialist's weights
$\theta_\tau=g_\phi(C_\tau)$ in one pass, with no test-time gradient steps and no
context in the query window.}
\label{fig:quadrant}
\end{figure}

We provide that map. We define a single formulation in which all four mechanisms
share a backbone and a budget (\S\ref{sec:method}, Fig.~\ref{fig:quadrant}) and run
the identical comparison across six tasks spanning five domains
(\S\ref{sec:exp}): few-shot sinusoid regression, parametric shape generation,
byte-level language modeling, meta-reinforcement learning, and clinical and genomic
few-shot classification. Rather than reporting only where emission wins, we
deliberately chart where it \emph{ties} and where it \emph{loses}, and we measure how
those boundaries move with scale.

The best-supported positive results are about \emph{cost at matched quality}:
emission ties the strongest amortized tabular baseline while producing a reusable
specialist rather than re-attending the support set per query (\S\ref{sec:tabular}),
and reaches noise-floor shape generation with a tiny per-instance program
(\S\ref{sec:shapes}). Its principal negative result is a language-modeling shortfall:
a one-pass emitted adapter recovers only a minority of the in-context performance
gain, and a rank sweep locates part of that shortfall in the emission channel's
capacity (\S\ref{sec:scaling}). A more speculative observation---that emitted
specialists compose in weight space (\S\ref{sec:compose}, Fig.~\ref{fig:morph})---
suggests the hypernetwork learns an approximately linear task-to-weight map, but we
report it with null controls and treat it as a direction rather than a headline.

\paragraph{Contributions.}
\begin{enumerate}[label=(C\arabic*),leftmargin=2.6em,topsep=2pt]
  \item A controlled four-way comparison of conditioning mechanisms across six tasks
        and five domains under matched budgets (\S\ref{sec:exp}).
  \item \textbf{Boundary maps} rather than wins: where emission dominates, ties, and
        loses---including a matched-budget scaling characterization on high-dimensional
        sequences, where emission recovers only a minority of the in-context gain that
        does not grow with scale, and a rank sweep locating part of the shortfall in
        the emission channel's capacity (\S\ref{sec:scaling}).
  \item \textbf{Mechanism ablations} establishing that the emitted specialist is
        genuinely task-conditioned, not a memorized prior (\S\ref{sec:ablation}).
  \item An \textbf{operationalized thesis}: a per-task ``resolution-$K$'' (the context
        size at which emission's error saturates) that predicts, a priori, where
        emission wins---turning ``low-dimensional, amortizable'' from a post-hoc label
        into a measured quantity (\S\ref{sec:resolution}).
  \item A \textbf{speculative observation}, reported with null controls: emitted
        specialists interpolate coherently in weight space, including on a function
        family not closed under addition (\S\ref{sec:compose}).
\end{enumerate}

\paragraph{Thesis (falsifiable).} \emph{Amortized weight-emission dominates within a
trained low-dimensional task family, at far lower test-time cost than gradient
adaptation or in-context attention, but it cannot match in-context attention for
high-dimensional sequence modeling, where it recovers only a minority of the
in-context gain---a shortfall that does not shrink with model scale over the range
we test.}

\section{Related Work}

\paragraph{Generating weights.} Hypernetworks~\citep{ha2016hypernetworks} introduced
networks that output the parameters of another network. Conditional Neural
Processes~\citep{garnelo2018cnp} and functa amortize this over a task distribution,
emitting an instance-specific function from a context set---the formulation our
shape and regression experiments instantiate. We adopt FiLM-style
modulation~\citep{perez2018film} as a low-dimensional alternative to emitting raw
weights.

\paragraph{Context-to-PEFT for language.} A line of work generates
parameter-efficient adapters from context: HyperTuning and HyperT5 emit soft
prefixes and LoRA weights from few-shot
demonstrations~\citep{phang2023hypertuning,phang2024gisting}, and more recent systems
map context to LoRA in a single pass~\citep{shine2026,hyperlora2024}. Our language
experiments use the same recipe; our contribution there is the controlled comparison
against in-context attention and the scaling trend (\S\ref{sec:scaling}), not the
adapter-generation mechanism.

\paragraph{Meta-learning and meta-RL.} Gradient-based meta-learning
(MAML~\citep{finn2017maml}, CAVIA~\citep{zintgraf2019cavia}) adapts via test-time
SGD; context-based meta-RL (PEARL~\citep{rakelly2019pearl}) conditions on an inferred
latent. \citet{beck2023hypernetworks} show hypernetwork-generated policies are
competitive with latent conditioning in meta-RL---a finding our RL experiment
reproduces and situates within the broader comparison.

\paragraph{Amortized tabular inference.} TabPFN~\citep{hollmann2023tabpfn} is a
pretrained transformer that classifies tabular data in-context in one forward pass;
it is the strongest baseline for our few-shot tabular tasks, and we compare against
it directly.

\paragraph{Weight-space arithmetic.} Task arithmetic~\citep{ilharco2023task} shows
that \emph{fine-tuned} weight deltas can be added and negated to compose model
behaviors. Our composition result (\S\ref{sec:compose}) is the hypernetwork
analogue: weights emitted in a single forward pass interpolate coherently, locating
the linearity in the learned task-to-weight map itself.

\paragraph{Decoupled training.} The appendix relates a negative result to synthetic
gradients / DNI~\citep{jaderberg2017dni} and decoupled greedy local
learning~\citep{nokland2019local,belilovsky2019greedy}.

\paragraph{Positioning.} We do not claim the weight-emission mechanism is new. Our
contribution is a controlled cross-domain map of \emph{when} it should be preferred
over the three alternatives, a scaling characterization of its principal failure
mode, and the observation that emitted specialists compose in weight space.

\section{Method: The MoM Quadrant}\label{sec:method}

\paragraph{Task and context.} A task $\tau$ is described by a context set
$C_\tau=\{(x_i,y_i)\}_{i=1}^K$ of $K$ labeled examples; methods are evaluated on
held-out queries $x_q$ drawn from the same $\tau$. We hold $K$, the query
distribution, and the specialist architecture fixed across all four mechanisms so
that differences reflect the conditioning mechanism rather than capacity or data.

\paragraph{The specialist and its emitter.} The specialist $f_\theta$ is a small
network (an MLP or a few-layer transformer, depending on the domain). An encoder
$g_\phi$ maps the context set to specialist parameters,
$\theta_\tau = g_\phi(C_\tau)$, and is trained end-to-end by differentiating the
query loss through the generated weights. We study two emission variants: a
\emph{direct} variant in which $g_\phi$ outputs the full weight vector $\theta_\tau$,
and a \emph{FiLM} variant in which it outputs per-layer scale and shift modulations
of a shared base specialist---trading expressiveness for far fewer emitted values.
The encoder is permutation-invariant in the context examples (a deep-sets pool or a
set transformer). Two implementation details proved necessary: emitting weights
against RMS-normalized targets (raw target magnitudes span orders of magnitude), and
matching the specialist's input featurization to the data's function class
(\S\ref{sec:exp}).

\paragraph{The four mechanisms.} Over this shared specialist (Fig.~\ref{fig:quadrant}):
\emph{zero-shot} uses fixed $\theta_0$ and ignores $C_\tau$; \emph{in-context}
prepends $C_\tau$ to the query and conditions through attention with no weight change;
\emph{gradient adaptation} produces $\theta_\tau$ by $n$ SGD steps on $C_\tau$
starting from a meta-learned $\theta_0$ (MAML) or from scratch (per-task fine-tune);
and \emph{MoM} sets $\theta_\tau = g_\phi(C_\tau)$ in a single forward pass.

\paragraph{Why emission can be cheaper at deployment.} MoM takes \emph{zero}
test-time gradient steps, unlike adaptation; the emitted specialist is typically
orders of magnitude smaller than the backbone and is \emph{reused} across all queries
for a task, unlike in-context attention, which re-processes the context for every
query and pays a context-length KV cache. These structural savings hold even when
accuracy is at parity, and are the lens through which we read the results.

\section{Experiments}\label{sec:exp}

We organize the six tasks from the most amortizable (low-dimensional task latent) to
the least. For each we give the protocol, the four-way comparison, and a verdict,
leading with where emission wins and stating plainly where it does not; where a
published protocol exists we reproduce its baselines to calibrate our harness.

\paragraph{What ``matched budget'' means.} Comparisons hold the specialist
architecture, the context size $K$, and the query distribution identical across
mechanisms, so that differences reflect the conditioning mechanism rather than
capacity or data. The one axis that is not automatically equal is \emph{meta-training
exposure}---how many task instances each learner sees before test time---because MAML
and MoM have different natural batch structures. Unless noted we report each method at
its own well-tuned training length (the operating point a practitioner would use), and
we additionally give a strictly \emph{episode-matched} sinusoid comparison
(\S\ref{sec:budget}) in which MoM and MAML see the same number of task episodes, to
show the advantage is not merely bought with extra data.

\subsection{Sinusoid few-shot regression (low-dim, amortizable)}\label{sec:sinusoid}
We follow the MAML/CAVIA protocol: each task is $y=A\sin(x-\varphi)$ with
$A\sim U[0.1,5]$, $\varphi\sim U[0,\pi]$, inputs in $[-5,5]$, $K{=}10$ context points,
and MSE over 100 held-out queries; the specialist is the standard 1--40--40--1 MLP
used by all methods. Our protocol-exact MAML reproduction ($0.171$ MSE) lands in the
published range ($0.19$--$0.29$), calibrating the harness.
\begin{table}[h]\centering\small
\begin{tabular}{lcc}
\toprule
Method & MSE $\downarrow$ & test-time grad steps \\
\midrule
CAVIA (published) & 0.19--0.21 & 5 \\
MAML (published) & 0.23--0.29 & 5 \\
MAML (our repro, protocol-exact) & $0.171\pm0.064$ & 5 \\
\textbf{MoM-direct} & $\mathbf{0.0004\pm0.0004}$ & \textbf{0} \\
\textbf{MoM-FiLM} & $\mathbf{0.0005\pm0.0001}$ & \textbf{0} \\
\bottomrule
\end{tabular}
\caption{Sinusoid 10-shot (MoM 5 seeds, MAML 3). At the compute point above, MoM's
mean MSE is $\sim$$2.5$ orders of magnitude below MAML's; under a \emph{budget-matched}
protocol (\S\ref{sec:budget}) the gap narrows to $\sim$$30\times$ but persists at zero
adaptation steps. OOD behavior is charted in \S\ref{sec:ood} (amplitude shift --- MoM
graceful, MAML diverges; frequency shift --- both fail, MAML $2\times$ more robust).}
\end{table}

\paragraph{How large is the gap, and how much is bought with compute?}\label{sec:budget}
Two caveats keep us from reading the raw ratio literally. First, the MoM mean
($0.0004$) carries a relative standard deviation near $100\%$ across seeds, so
quoting a two-significant-figure ratio ($0.171/0.0004$) overstates the precision;
the honest statement is that emission lands $2$--$3$ orders of magnitude below
gradient adaptation here. Second, the operating-point MoM above sees $\sim$$4\times$
more task episodes than our MAML ($7.7$M vs $1.75$M), so part of the gap is bought
with data. We therefore retrain MoM at a strictly episode-matched budget---the same
$1.75$M task episodes MAML sees. The advantage shrinks but does not vanish:
MoM-direct $0.0050\pm0.0006$ ($\sim$$34\times$ below MAML's $0.171$) and MoM-FiLM
$0.0073\pm0.0009$ ($\sim$$23\times$), still at \emph{zero} test-time gradient steps.
We read this not as ``emission dominates'' in the abstract but as a concrete,
interpretable fact: on a family whose task latent is two-dimensional, a set encoder
that meta-learns the near-closed-form context$\to$specialist map beats adapting a
generic initialization by SGD, and it does so with an order-of-magnitude margin that
is only partly attributable to training data. The mechanism ablations
(\S\ref{sec:ablation}) confirm the encoder is doing inference, not lookup: query
error falls monotonically with context size and resolves precisely as $K$ passes the
two latent parameters.

\subsection{Vector-shape generation (low-dim, amortizable)}\label{sec:shapes}
Each task is a parametric 2D curve (circle, square, spiral, etc.) under random
rotation, scale, and noise, observed as a sequence of points. The natural baseline is
an autoregressive transformer that rolls out the curve point by point; MoM emits a
specialist $t\mapsto(x,y)$ from a few context points and renders the whole curve in
parallel. Accuracy is comparable across the parallel methods; the interesting
question is what the resource gap actually buys, so we decompose it.

To separate \emph{parallel decoding} from \emph{weight emission}, we add a
non-autoregressive baseline: a single generator that reads the context and predicts
all points in one parallel pass, with no hypernetwork bottleneck. It isolates the two
effects (Table~\ref{tab:shapes}). Most of the end-to-end speedup is architectural:
moving from a serial rollout to a parallel decoder is a $\sim$$20\times$ win by itself,
and a non-AR generator already reaches the noise floor. Emission then contributes a
further $\sim$$3\times$ (the emitted specialist is much smaller than the monolithic
parallel net, so its forward pass is cheaper), and---the emission-specific benefit
that no other method has---the specialist is a tiny \emph{reusable} program
(1.7k floats, or 132 for the FiLM variant) that can be cached and re-run at
$\sim$$0.9$\,ms per 256 shapes without re-reading the context. We therefore do not
credit the headline $60\times$ to emission; the honest reading is a large parallelism
win, a modest emission-size win, and a genuinely small and reusable per-instance
program.
\begin{table}[h]\centering\small
\begin{tabular}{lccc}
\toprule
Method & continuation MSE $\downarrow$ & gen time (256 shapes) & per-instance size \\
\midrule
AR transformer (serial) & 0.0041 & 625 ms & 202k params \\
non-AR generator (parallel, no emission) & 0.0021 & 31 ms & 319k params \\
\textbf{MoM hyper$\to$specialist} & \textbf{0.0018} (noise floor) & \textbf{10 ms} & 1.7k params \\
\quad specialist alone (amortized) & 0.0018 & \textbf{0.9 ms} & \textbf{132--1.7k floats} \\
\bottomrule
\end{tabular}
\caption{Decomposing the shape-generation speedup. AR$\to$non-AR is $\sim$$20\times$
(parallel vs.\ serial decoding, an architecture effect); non-AR$\to$MoM is a further
$\sim$$3\times$ (the emitted specialist is smaller than the monolithic parallel net).
Accuracy is at the noise floor for both parallel methods. Emission's distinctive
contribution is not raw speed but the tiny \emph{reusable} per-instance program.
\emph{Gotcha for appendix: specialist function-class must match data---a periodic
Fourier basis fails on open curves; diagnose by direct per-instance fitting.}}
\label{tab:shapes}
\end{table}

\subsection{Few-shot tabular: clinical \& genomic identification}\label{sec:tabular}
We use two real datasets: the UCI HCV liver panel (12 blood markers; healthy vs.\
liver disease) and the TCGA PAN-CAN RNA-seq corpus (top-1000 high-variance genes; 5
tumor types). On the full data our logistic-regression and random-forest references
match published SOTA ($0.99$ AUC; $0.999$ accuracy), confirming the data is solved
and calibrating the harness. The MoM-relevant regime is therefore few-shot: a task is
a small balanced support set, and we compare emission against in-context attention,
per-task gradient fitting, and the state-of-the-art amortized tabular model TabPFN.
\begin{table}[h]\centering\small
\begin{tabular}{lccc}
\toprule
& \multicolumn{2}{c}{Blood (HCV), 16-shot} & Genes (TCGA), 5-shot \\
\cmidrule(lr){2-3}\cmidrule(lr){4-4}
Method & ROC-AUC $\uparrow$ & infer cost & accuracy $\uparrow$ \\
\midrule
Full-data SOTA (ref) & 0.989--0.994 & --- & 0.999 \\
TabPFN (SOTA amortized) & 0.967 & 601 ms/task, re-attend/query & 0.991$^\dagger$ \\
\textbf{MoM (emit classifier)} & \textbf{0.966} & reusable specialist, $\sim$ms & 0.998 \\
in-context attention & 0.925 & full support/query & 1.000 \\
per-task logistic fit & 0.930 & grad steps/task & 0.996 \\
\bottomrule
\end{tabular}
\caption{MoM \textbf{ties SOTA TabPFN} on blood (0.966 vs 0.967) while beating
gradient-tuning and in-context --- and emits a \emph{reusable} specialist whereas
TabPFN re-attends the full support set for every query. Genes is a saturated tie
(near-separable). $^\dagger$TabPFN capped at PCA-20 by its feature limit; MoM uses
all 1000 genes.}
\end{table}

\subsection{Meta-RL: HalfCheetah-Vel (PEARL protocol)}\label{sec:metarl}
Tasks are target running velocities; an episode-1 trajectory is the context for
episode-2 control. We hold the SAC machinery and a context encoder fixed and vary
only the conditioning mechanism: \emph{latent} concatenates the inferred task
embedding to the policy input (PEARL-style), while \emph{MoM} uses it to FiLM-modulate
the policy's weights. We report the standard best-checkpoint return and the final
return over 3 seeds.
\begin{table}[h]\centering\small
\begin{tabular}{lccc}
\toprule
Method & best-ckpt return $\uparrow$ & final return & conditioning \\
\midrule
PEARL (published) & $\approx-65$ & --- & latent concat \\
latent (our PEARL-style) & $-50\pm15$ & $-48\pm11$ & concat \\
\textbf{MoM (FiLM policy)} & $\mathbf{-57\pm5}$ & $-70\pm11$ & emit weights \\
\bottomrule
\end{tabular}
\caption{3 seeds. On the standard best-checkpoint metric, latent and MoM are a
\textbf{statistical tie} (overlapping CIs); MoM is markedly \textbf{more consistent}
(std 5 vs 15). Latent edges MoM on final return. Both reach/beat published PEARL.
Consistent with Beck et al.\ (2023): hypernet policies competitive, not dominant.}
\end{table}

\subsection{Language modeling (high-dim): enwik8 byte LM}\label{sec:lm}
The backbone is a frozen byte-level transformer pretrained on enwik8. MoM's
hyper-LoRA reads 512 context bytes and emits a rank-4 LoRA adapter; the model then
predicts a contiguous following chunk with no context in its own window. We compare
against zero-shot, in-context (the 512 bytes prepended), and a per-document
Adam-tuned LoRA. This is a high-dimensional task latent (the ``style'' of an
arbitrary document), and emission does not win.
\begin{table}[h]\centering\small
\begin{tabular}{lcc}
\toprule
Method & bits/char $\downarrow$ & context tokens in window \\
\midrule
zero-shot (frozen 5M base) & 1.814 & 0 \\
in-context (512 prepended) & \textbf{1.645} & 512 \\
\textbf{MoM hyper-LoRA} & 1.789 & \textbf{0} \\
per-doc Adam-tuned LoRA & 2.557 & 0 \\
\bottomrule
\end{tabular}
\caption{MoM captures only $\sim$15\% of the in-context gain at this base
($14.0\pm0.9\%$ under the matched-budget $3$-seed protocol of \S\ref{sec:scaling}), but
beats gradient tuning, which is \emph{worse than zero-shot}. See \S\ref{sec:scaling}
for the rank sweep and matched-budget scaling.}
\end{table}

\subsection{Puzzle solving: ARC (high-dim, compositional)}\label{sec:arc}
We train on procedurally generated RE-ARC tasks: MoM reads three input--output grid
pairs and emits a specialist that transforms a held-out test grid. We evaluate on
held-out instances of seen rules and on the real ARC training and evaluation sets.
\begin{itemize}[leftmargin=1.4em]
  \item Held-out generated (seen rules, new instances): \textbf{11.5\%} exact match.
  \item Real ARC training grids: 3.6\%. \quad Real ARC \textbf{evaluation (unseen
        rules): 0.0\%}.
  \item \textbf{Verdict:} clean negative. Amortization \emph{interpolates the trained
        task family} but does not \emph{synthesize new programs}; ARC SOTA (40--55\%)
        uses search/LLMs far outside our budget.
\end{itemize}

\section{Analysis}

\subsection{Mechanism ablations: is the specialist really task-conditioned?}\label{sec:ablation}
\begin{figure}[t]\centering
\includegraphics[width=\textwidth]{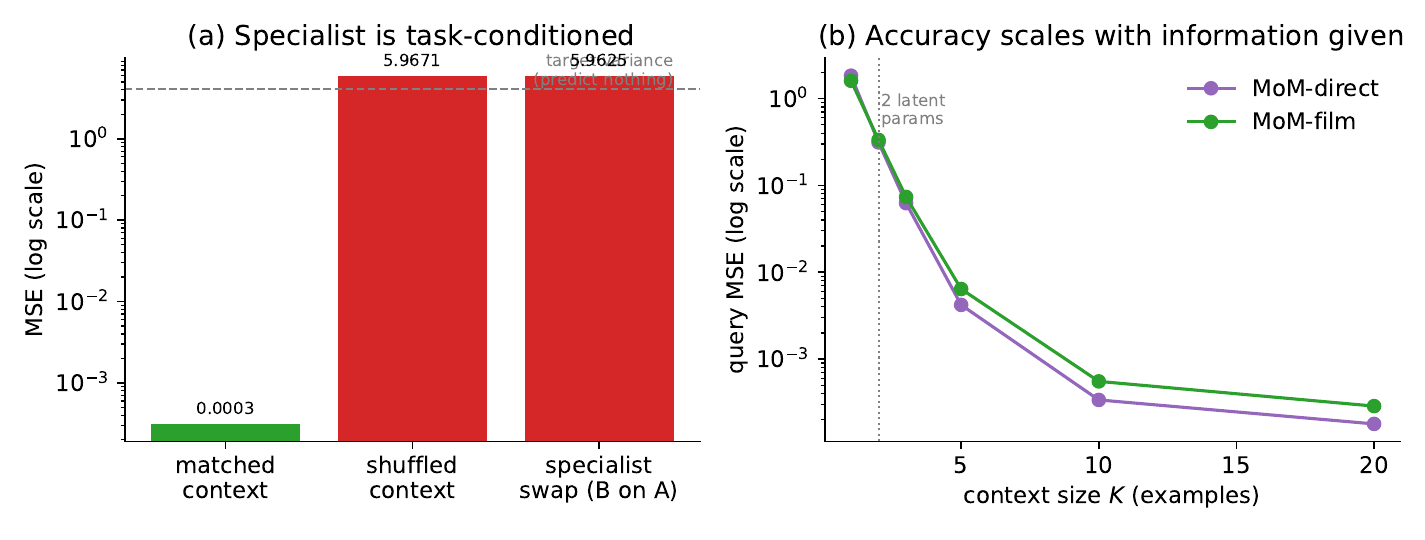}
\caption{Mechanism ablations on the sinusoid MoM. (a) Replacing the context with
another task's examples, or running task $A$'s specialist on task $B$, raises error
$\sim$19{,}000$\times$---past the variance of predicting nothing---so the specialist
is fully context-determined. (b) Query error falls monotonically with context size
and resolves once $K$ exceeds the two latent task parameters: the encoder performs
inference, not retrieval.}
\label{fig:ablation}
\end{figure}

A natural objection is that the encoder ignores its context and the system merely
memorizes the task distribution. Three ablations on the sinusoid MoM refute this.
\begin{itemize}[leftmargin=1.4em]
  \item \textbf{Context-shuffle:} matched-context MSE 0.0003 vs wrong-task-context
        5.97 ($\sim$19{,}000$\times$ worse, and worse than the 4.14 target variance ---
        confidently wrong, not a memorized prior).
  \item \textbf{Specialist-swap:} task-A specialist on A 0.0003 vs B-specialist on A
        5.96. Each emitted network is genuinely task-specific.
  \item \textbf{Context-size sweep:} $K{=}1{:}1.84,\,2{:}0.31,\,3{:}0.06,\,5{:}0.004,
        \,10{:}0.0003$ --- monotonic, resolves as $K$ passes the 2 latent params.
        Inference, not lookup (Fig.~\ref{fig:ablation}b).
\end{itemize}

\subsection{Operationalizing the thesis: resolution-$K$}\label{sec:resolution}
The thesis predicts wins for ``low-dimensional, amortizable'' families, but that
predicate is only useful if it can be measured \emph{before} seeing which method won.
The ablation of \S\ref{sec:ablation} suggests an instrument: query error falls with
context size and saturates once $K$ passes the task's intrinsic dimension. We turn
this into a per-task number, \emph{resolution-$K$}: the smallest context size at which
MoM's error reaches within $10\%$ of its total drop from $K{=}1$ (or, for tasks
already solved at the smallest tested $K$, that smallest $K$). It is estimated purely
from the emission method's own $K$-sweep, with no reference to any baseline.
\begin{table}[h]\centering\small
\begin{tabular}{lccl}
\toprule
task & resolution-$K$ & nominal feature dim & emission outcome \\
\midrule
genes (TCGA) & 1 & $\ge$1000 & saturated tie \\
blood (HCV) & 2 & 12 & \textbf{ties SOTA} (TabPFN) \\
sinusoid & 3 & 2 & \textbf{wins} ($30$--$400\times$ vs MAML) \\
shapes & 12 & $\sim$6 & \textbf{wins} (noise floor, cheaper) \\
enwik8 LM & \emph{unresolved} ($>512$B) & very high & \textbf{loses} to in-context \\
ARC (unseen) & \emph{unresolved} & very high & \textbf{loses} ($0\%$) \\
\bottomrule
\end{tabular}
\caption{Resolution-$K$ predicts the outcome a priori: emission ties-or-wins exactly
on the tasks that resolve at a small \emph{finite} $K$, and loses on the two tasks
whose error never saturates within the available context. Note that resolution-$K$
tracks intrinsic \emph{task} complexity, not nominal feature count---the genomic task
has $\ge$1000 features but is pinned by a single shot per class, and MoM ties there;
enwik8 has a small alphabet but an unbounded document-style latent, and MoM loses.
This is the measurable content of ``amortizable.''}
\label{tab:resolution}
\end{table}

\subsection{Out-of-distribution behavior: two-sided}\label{sec:ood}
The sinusoid family lets us probe the boundary of amortization directly, by shifting
the test task off the training distribution along two axes. The two shifts move the
boundary in opposite directions, which is the point.

\emph{Amplitude shift (a $y$-scaling of the same function class).} As we push the
amplitude $A$ past its training range $[0.1,5]$, MoM degrades gracefully---about
$10\times$ per band while staying $30$--$250\times$ ahead of MAML---whereas MAML's
inner loop \emph{diverges} at far-OOD (its large-amplitude error explodes the fixed
inner-lr updates, reaching NaN at $20$ steps). Amortization extrapolates a scaling
shift; gradient adaptation destabilizes on it.

\emph{Frequency shift (a structural change to the function class).} As we raise the
angular frequency $\omega$ above the trained $\omega{=}1$, the ranking inverts: MoM
breaks hard (error exceeds the task variance $\approx4$, i.e.\ worse than predicting
zero), while MAML degrades roughly $2\times$ less. Neither is usable in absolute
terms, but gradient adaptation keeps a relative edge under structural novelty.
\begin{table}[h]\centering\small
\begin{tabular}{lcccc}
\toprule
& \multicolumn{2}{c}{Amplitude shift (MSE $\downarrow$)} & \multicolumn{2}{c}{Frequency shift (MSE $\downarrow$)} \\
\cmidrule(lr){2-3}\cmidrule(lr){4-5}
band & MoM-direct & MAML ($5$/$20$ step) & band & MoM-direct \,/\, MAML$_5$ \\
\midrule
in-dist $A{\in}[0.1,5]$ & 0.0003 & 0.125 / 0.092 & $\omega{=}1$ & 0.0003 / 0.119 \\
near $A{\in}[5,6]$ & 0.0030 & 0.744 / 0.516 & $\omega{\in}[1.2,1.5]$ & 3.80 / 1.78 \\
far $A{\in}[6,7]$ & 0.0232 & 1.859 / \textbf{NaN} & $\omega{\in}[1.5,2]$ & 8.22 / 3.74 \\
\bottomrule
\end{tabular}
\caption{OOD probes on sinusoid. \textbf{Scaling shift} (amplitude): MoM extrapolates
gracefully, MAML diverges. \textbf{Structural shift} (frequency): both fail, MAML
$\sim2\times$ more robust. Amortization's advantage is confined to its trained
function class and to scaling extrapolation within it---and its failure mode is
confident wrongness, not abstention.}
\label{tab:ood}
\end{table}

\subsection{Weight-space composition (a speculative observation)}\label{sec:compose}
\begin{figure}[h]\centering
\includegraphics[width=0.95\textwidth]{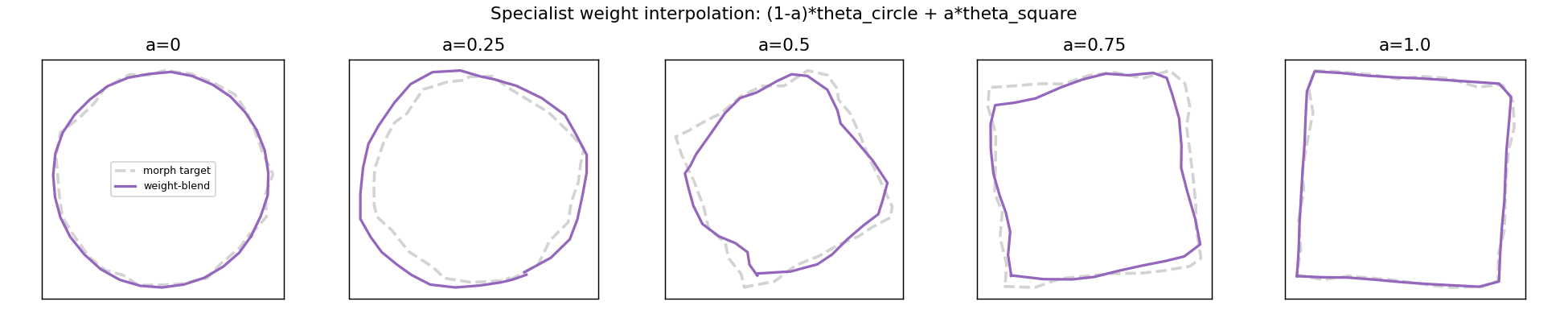}
\caption{Specialist weight interpolation on the (non-closed) shapes family. Purple:
output of the blended specialist $(1{-}\alpha)\theta_{\text{circle}}+\alpha
\theta_{\text{square}}$; gray dashed: the geometric morph target. The blend tracks the
specific morph ($5.3\times$ closer to it than to a mismatched pair's morph) and is
genuinely intermediate rather than snapping to an endpoint.}
\label{fig:morph}
\end{figure}

If the encoder maps tasks to weights approximately linearly, interpolating two
emitted specialists should track the corresponding blend of their functions. It does,
on our low-dimensional synthetic families. Interpolating
$\theta=(1{-}\alpha)\theta_A+\alpha\theta_B$ yields a midpoint output that tracks the
geometric morph of the two shapes: on the \textbf{shapes} family (which is \emph{not}
closed under addition) the midpoint MSE to the true morph is $0.061$, versus a
$1.45$ mean shape distance---a coherent circle$\to$square interpolation
(Fig.~\ref{fig:morph}).

\paragraph{Is this real, or just a flat basin?} A coherent midpoint could be an
artifact: if every specialist in this region produced similar outputs, any blend
would look smooth. Two null controls argue otherwise. \emph{Specificity:} the
midpoint's error against \emph{this} pair's morph ($0.061$) is $5.3\times$ smaller
than against a \emph{different} random pair's morph ($0.323$), so the blend tracks
the specific geometric target rather than a generic average. \emph{Non-degeneracy:}
the midpoint sits far from either endpoint specialist's output ($0.42$ to each), so
it is genuinely intermediate, not collapsed onto one mode; and the error along the
interpolation path is smooth and symmetric ($0.003\!\to\!0.031\!\to\!0.061\!\to\!
0.031\!\to\!0.003$ across $\alpha$). The closest prior relative is task
arithmetic~\citep{ilharco2023task}, which composes \emph{fine-tuned} weight deltas;
here composition holds for weights produced in a single forward pass. We flag two
honest limits: this is demonstrated only on synthetic $2$--$6$-parameter families,
and---given the $0.0\%$ ARC result on unseen rules (\S\ref{sec:arc})---it is
interpolation \emph{within} the trained task manifold, not synthesis of genuinely new
programs. Whether emitted specialists for real tasks compose, and whether arithmetic
($A{-}B{+}C$) reaches unseen combinations, is future work.

\subsection{Scaling, and the capacity of the emission channel}\label{sec:scaling}
On language modeling a one-pass emitted adapter recovers only a \emph{minority} of the
in-context gain, at every scale we tested. We re-ran the scaling sweep with matched
budgets---identical $8$k-step pre-training across all sizes and three seeds per
point---so the numbers are no longer confounded by the training-length mismatch of an
earlier single-seed sweep. The captured fraction is $14.0\pm0.9\%$ at a $5$M base and
$11.2\pm0.5\%$ at $15$M (Table~\ref{tab:scaling}); the $1$M base is degenerate
(in-context gain only $0.04$, ratio meaningless) and is excluded. Two observations
survive the tighter protocol. First, the decrease from $5$M to $15$M is now real
rather than an artifact: the seed intervals do not overlap, and---unlike the earlier
sweep, where the in-context gain appeared to grow---the in-context gain here is
essentially flat ($0.182$ vs $0.184$) across the two sizes, so the shrinking fraction
is not being manufactured by a growing denominator. Second, and this is the honest
limit, two usable points still cannot establish a \emph{frontier} trend: whether the
fraction keeps shrinking or saturates beyond $15$M is unresolved, and we do not claim
otherwise. The defensible claim is the one the thesis needs: \emph{for high-dimensional
sequence modeling, emission recovers only a minority of the in-context gain, does not
close it, and does not grow its share over the range we can afford to test.}

\paragraph{Is the shortfall capacity or dimensionality?} A natural worry is that we
have conflated two different failures under one ``high-dimensional'' label: the
language shortfall could be an \emph{emission-channel capacity} limit ($512$ context
bytes squeezed into a rank-$4$ LoRA), which more rank would relieve, or an intrinsic
\emph{dimensionality} limit, which it would not. Sweeping the LoRA rank on the fixed
$5$M base separates them (Table~\ref{tab:rank}). Capture rises monotonically with
rank---$5.2\%,\,13.1\%,\,18.1\%,\,21.0\%$ at ranks $1,4,16,64$---so capacity \emph{is}
implicated: a wider adapter genuinely recovers more of the in-context gain. But the
returns diminish sharply (rank $64$ quadruples rank $16$'s parameters for a further
$3$ points) and the fraction plateaus far below $100\%$, so capacity is not the whole
story---even a generously-ranked one-pass adapter leaves most of the gap to attention
unrecovered. This is a partial-capacity bottleneck, and it is qualitatively different
from the ARC failure (\S\ref{sec:arc}; $0\%$ on unseen rules), a compositional-synthesis
wall that no adapter rank addresses.
\begin{table}[h]\centering\small
\begin{minipage}[t]{0.52\textwidth}\centering
\begin{tabular}{lcccc}
\toprule
base & zs & ic & hl & \% ICL gain \\
\midrule
1.0M$^\ddagger$ & 2.556 & 2.515 & 2.487 & degen. \\
5.1M & 1.831 & 1.648 & 1.796 & $14.0{\pm}0.9$ \\
14.7M & 1.652 & 1.468 & 1.620 & $11.2{\pm}0.5$ \\
\bottomrule
\end{tabular}
\captionof{table}{Scaling, matched $8$k pre-train, $3$ seeds. Capture is a minority at
both usable scales and does not grow. $^\ddagger$$1$M degenerate (ICL gain $0.04$).}
\label{tab:scaling}
\end{minipage}\hfill
\begin{minipage}[t]{0.44\textwidth}\centering
\begin{tabular}{lccc}
\toprule
rank & params & hl & \% ICL gain \\
\midrule
1 & 10.8k & 1.812 & 5.2 \\
4 & 43k & 1.798 & 13.1 \\
16 & 172k & 1.789 & 18.1 \\
64 & 688k & 1.784 & 21.0 \\
\bottomrule
\end{tabular}
\captionof{table}{LoRA-rank sweep, fixed $5$M base (zs $1.821$, ic $1.643$). Capture
rises then plateaus $\ll100\%$: partial capacity limit.}
\label{tab:rank}
\end{minipage}
\end{table}

\section{Limitations}
Our experiments are at small scale---5--15M-parameter language models, MLP and
few-layer-transformer specialists, two consumer GPUs. We measure a scaling
\emph{trend}, but the absolute regime is small, and the language and ARC conclusions
should be read as bounds at this scale rather than at frontier scale. Emission fails
on high-dimensional and compositional tasks (language modeling, ARC), and on language
the failure \emph{grows} with scale, so the method is not a general-purpose substitute
for attention. Out of distribution, amortization extrapolates gracefully under
scaling shifts but breaks under structural shifts and is confidently wrong rather than
abstaining---a safety-relevant failure mode for deployment. Finally, weight-space
composition is demonstrated on synthetic families; whether emitted specialists for
real tasks compose as cleanly is open.

\section{Conclusion}
Across six tasks and five domains we find a consistent pattern: emitting a
task-specific specialist dominates when the task family is low-dimensional and
amortizable---matching or beating gradient adaptation and the strongest amortized
baselines at far lower test-time cost---but cannot match in-context attention for
high-dimensional sequence modeling, where a one-pass adapter recovers only a minority
of the in-context gain (a partial capacity limit) that does not close with scale. The practical
decision rule that follows is simple: \emph{emit a specialist when the task family is
narrow and deployment cost matters; attend in-context when the task latent is rich,
especially at scale.} Two further findings sharpen the picture: mechanism ablations
show the emitted specialist is genuinely task-determined, and emitted specialists
compose algebraically in weight space---a property absent from the alternatives and a
promising direction for turning amortized emission from a conditioning trick into a
manipulable representation of tasks.

\appendix
\section{Negative result: predicted gradients vs.\ local losses}
\label{app:negative}
This project began with a different question---whether a transformer could be trained
without global backpropagation by having each layer \emph{predict} its own incoming
gradient (a decoupled-neural-interfaces / synthetic-gradient
scheme~\citep{jaderberg2017dni})---and the negative result that emerged is worth
recording, both because it is clean and because it motivated the amortization view
that became MoM.

On a shape-classification task, per-token gradient predictors with RMS-normalized
targets and a bootstrapped target chain do train a small transformer with zero global
backward passes, reaching $96$--$99.6\%$ accuracy at two layers. But the approach has
two limits. First, it does not scale in depth: bootstrap error compounds and accuracy
collapses beyond a few layers \emph{unless} each stage is given an auxiliary local
loss that supplies an exact gradient anchor. Second---and this is the negative
result---once those local anchors are present, the gradient predictors are
\emph{redundant}: an ablation that removes them and keeps only the per-stage auxiliary
losses (greedy local learning,~\citealp{nokland2019local,belilovsky2019greedy})
matches end-to-end backpropagation ($100\%$ / $99.8\%$ at 4 / 6 layers on
classification; $0.0033$ vs.\ $0.0022$ MSE on autoregressive generation) while holding
peak training memory nearly flat in depth ($740$\,MB vs.\ backpropagation's $6927$\,MB
at 24 layers, a $9.4\times$ reduction).

A useful diagnostic emerged along the way: \emph{gradient predictability decays over
training}. The cosine similarity between a layer's true and predicted gradient is
$\approx0.99$ at initialization but drops sharply to roughly $0.2$--$0.5$ once the
loss converges, as gradients grow noisier and less structured. This weakens any
synthetic-gradient method in the late, low-loss regime and helps explain why such
methods help least exactly when one most wants a shortcut. The broader lesson---that
\emph{amortizing} a learned signal beats predicting a moving target---carried directly
into the specialist-emission framing of the main paper. This appendix could stand as a
short negative-results contribution on its own.

\section{Reproducibility}
All datasets are public: UCI HCV, UCI TCGA PAN-CAN RNA-seq, enwik8, ARC-AGI with the
RE-ARC procedural generators, and Gymnasium HalfCheetah-v5. All models are implemented
in PyTorch and trained on consumer GPUs (2$\times$NVIDIA~A2 and 2$\times$RTX~5070).
Code and configurations: \url{https://github.com/johnchowell/Model-of-Models}. Per-experiment settings are in
Table~\ref{tab:hparams}; results report multiple seeds where noted in
\S\ref{sec:exp} (sinusoid: 5; meta-RL: 3; tabular: 5-fold cross-validation).

\begin{table}[h]\centering\small
\setlength{\tabcolsep}{4pt}
\begin{tabular}{lp{0.62\textwidth}}
\toprule
Experiment & Key settings \\
\midrule
Sinusoid (\S\ref{sec:sinusoid}) & specialist MLP 1--40--40--1; encoder deep-sets, $d{=}128$; MoM
  $30$k steps, batch $256$, Adam lr $10^{-3}$ cosine; MAML $70$k meta-steps, batch
  $25$, inner lr $0.01$ ($1$ step), outer lr $10^{-3}$, 2nd-order; $K{=}10$. \\
Shapes (\S\ref{sec:shapes}) & specialist Fourier($t$)$+t$ $\to$ MLP, $1.7$k params (FiLM:
  $132$ mods); encoder transformer $d{=}128$; $8$k steps, batch $256$, lr $10^{-3}$;
  context $12$ points, curve length $32$. \\
Tabular (\S\ref{sec:tabular}) & few-shot net $d{=}128$ (genes) / row-encoder + FiLM head; $3$--$4$k
  steps, batch $32$--$64$, lr $10^{-3}$, wd $10^{-4}$; blood $K{=}16$ balanced,
  genes 5-way 5-shot, top-$1000$ genes; TabPFN v2 (PCA-$20$ for genes). \\
Meta-RL (\S\ref{sec:metarl}) & SAC, policy/critic hidden $256$; $300$ iters, batch $256$, lr
  $3{\times}10^{-4}$, $\gamma{=}0.99$, $\tau{=}5{\times}10^{-3}$; context encoder
  $z{\in}\mathbb{R}^{16}$, $64$ context transitions; episodes $200$ steps. \\
enwik8 (\S\ref{sec:lm}) & base byte-LM $d{=}256$, $6$ layers, $T{=}768$, pretrain $8$k steps,
  batch $32$, lr $3{\times}10^{-4}$ OneCycle; hyper-LoRA rank $4$ on
  $\{o,\text{fc}_2\}$, $4$k steps; context $512$ bytes, target $256$. \\
ARC (\S\ref{sec:arc}) & grid transformer, enc $4$ / dec $6$ layers, $d{=}256$, $12{\times}12$
  canvas; $60$k steps, batch $32$, lr $3{\times}10^{-4}$; $3$ example pairs; trained
  on RE-ARC, grids $\le 12{\times}12$. \\
Scaling (\S\ref{sec:scaling}) & bases $(d,L)\in\{(128,4),(256,6),(384,8)\}$; pretrain $8$k
  (matched to the headline base), hyper-LoRA $4$k steps each, $3$ seeds. \\
\bottomrule
\end{tabular}
\caption{Per-experiment configurations.}
\label{tab:hparams}
\end{table}

\bibliographystyle{plainnat}
\bibliography{references}

@inproceedings{ha2016hypernetworks,
  title={HyperNetworks},
  author={Ha, David and Dai, Andrew and Le, Quoc V.},
  booktitle={International Conference on Learning Representations (ICLR)},
  year={2017}
}

@inproceedings{garnelo2018cnp,
  title={Conditional Neural Processes},
  author={Garnelo, Marta and Rosenbaum, Dan and Maddison, Chris J. and Ramalho,
          Tiago and Saxton, David and Shanahan, Murray and Teh, Yee Whye and
          Rezende, Danilo J. and Eslami, S. M. Ali},
  booktitle={International Conference on Machine Learning (ICML)},
  year={2018}
}

@inproceedings{perez2018film,
  title={FiLM: Visual Reasoning with a General Conditioning Layer},
  author={Perez, Ethan and Strub, Florian and de Vries, Harm and Dumoulin, Vincent
          and Courville, Aaron},
  booktitle={AAAI Conference on Artificial Intelligence},
  year={2018}
}

@inproceedings{phang2023hypertuning,
  title={HyperTuning: Toward Adapting Large Language Models without Back-propagation},
  author={Phang, Jason and Mao, Yi and He, Pengcheng and Chen, Weizhu},
  booktitle={International Conference on Machine Learning (ICML)},
  year={2023}
}

@article{phang2024gisting,
  title={Investigating the Effectiveness of HyperTuning via Gisting},
  author={Phang, Jason},
  journal={arXiv preprint arXiv:2402.16817},
  year={2024}
}

@article{shine2026,
  title={{SHINE}: A Scalable In-Context Hypernetwork for Mapping Context to {LoRA}
         in a Single Pass},
  author={Liu, Yewei and Wang, Xiyuan and Mao, Yansheng and Gelbery, Yoav and
          Maron, Haggai and Zhang, Muhan},
  journal={arXiv preprint arXiv:2602.06358},
  year={2026}
}

@inproceedings{hyperlora2024,
  title={{HyperLoRA}: Efficient Cross-task Generalization via Constrained Low-Rank
         Adapters Generation},
  author={Lv, Chuancheng and Li, Lei and Zhang, Shitou and Chen, Gang and Qi,
          Fanchao and Zhang, Ningyu and Zheng, Hai-Tao},
  booktitle={Findings of the Association for Computational Linguistics: EMNLP 2024},
  pages={16376--16393},
  year={2024}
}

@inproceedings{ilharco2023task,
  title={Editing Models with Task Arithmetic},
  author={Ilharco, Gabriel and Ribeiro, Marco Tulio and Wortsman, Mitchell and
          Gururangan, Suchin and Schmidt, Ludwig and Hajishirzi, Hannaneh and
          Farhadi, Ali},
  booktitle={International Conference on Learning Representations (ICLR)},
  year={2023}
}

@inproceedings{finn2017maml,
  title={Model-Agnostic Meta-Learning for Fast Adaptation of Deep Networks},
  author={Finn, Chelsea and Abbeel, Pieter and Levine, Sergey},
  booktitle={International Conference on Machine Learning (ICML)},
  year={2017}
}

@inproceedings{zintgraf2019cavia,
  title={Fast Context Adaptation via Meta-Learning},
  author={Zintgraf, Luisa and Shiarlis, Kyriacos and Kurin, Vitaly and Hofmann,
          Katja and Whiteson, Shimon},
  booktitle={International Conference on Machine Learning (ICML)},
  year={2019}
}

@inproceedings{rakelly2019pearl,
  title={Efficient Off-Policy Meta-Reinforcement Learning via Probabilistic Context
         Variables},
  author={Rakelly, Kate and Zhou, Aurick and Finn, Chelsea and Levine, Sergey and
          Quillen, Deirdre},
  booktitle={International Conference on Machine Learning (ICML)},
  year={2019}
}

@article{beck2023hypernetworks,
  title={Hypernetworks in Meta-Reinforcement Learning},
  author={Beck, Jacob and Jackson, Matthew and Vuorio, Risto and Whiteson, Shimon},
  journal={Conference on Robot Learning (CoRL)},
  year={2023}
}

@inproceedings{hollmann2023tabpfn,
  title={{TabPFN}: A Transformer That Solves Small Tabular Classification Problems
         in a Second},
  author={Hollmann, Noah and M{\"u}ller, Samuel and Eggensperger, Katharina and
          Hutter, Frank},
  booktitle={International Conference on Learning Representations (ICLR)},
  year={2023}
}

@article{jaderberg2017dni,
  title={Decoupled Neural Interfaces using Synthetic Gradients},
  author={Jaderberg, Max and Czarnecki, Wojciech Marian and Osindero, Simon and
          Vinyals, Oriol and Graves, Alex and Silver, David and Kavukcuoglu, Koray},
  journal={International Conference on Machine Learning (ICML)},
  year={2017}
}

@inproceedings{nokland2019local,
  title={Training Neural Networks with Local Error Signals},
  author={N{\o}kland, Arild and Eidnes, Lars Hiller},
  booktitle={International Conference on Machine Learning (ICML)},
  year={2019}
}

@inproceedings{belilovsky2019greedy,
  title={Greedy Layerwise Learning Can Scale to {ImageNet}},
  author={Belilovsky, Eugene and Eickenberg, Michael and Oyallon, Edouard},
  booktitle={International Conference on Machine Learning (ICML)},
  year={2019}
}

\end{document}